\documentclass[a4paper]{article}
\usepackage[preprint]{neurips_2024}
\usepackage{algorithm}
\usepackage{algorithmicx}
\usepackage{algpseudocode}
\usepackage{amsmath}
\usepackage{setspace}
\usepackage{hyperref}
\usepackage{multirow}
\usepackage{booktabs}
\usepackage{makecell}
\usepackage{graphicx} 
\usepackage{wrapfig} 
\usepackage{caption}

\title{Motif 2 12.7B technical report}
\author{Motif Technologies}

\begin{document}

\maketitle

\begin{center}
    \includegraphics[width=0.23\textwidth]{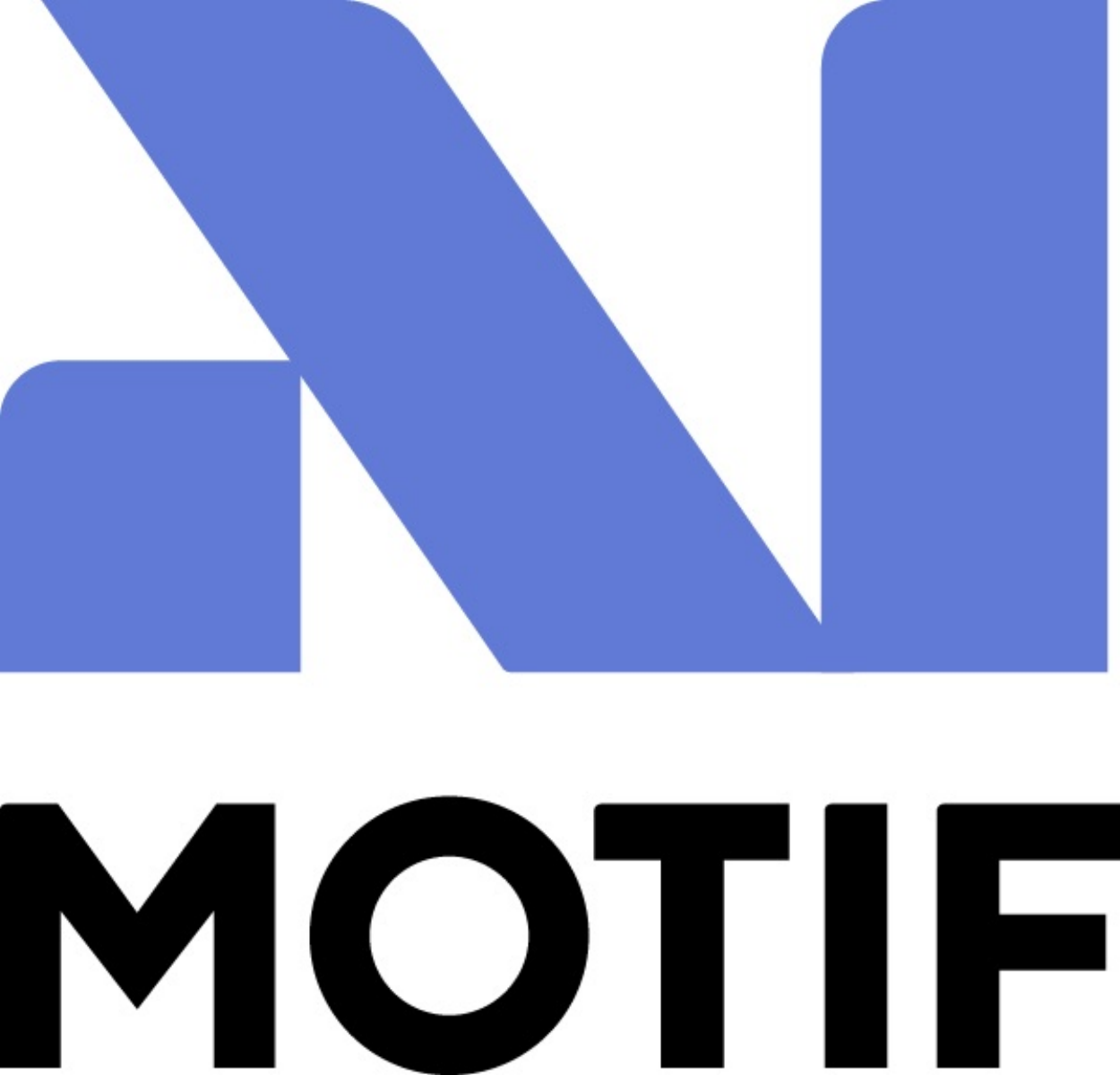} 
\end{center}

\vspace*{1cm}

\begin{abstract}

We introduce Motif-2-12.7B, a new open-weight foundation model that pushes the efficiency frontier of large language models by combining architectural innovation with system-level optimization. Designed for scalable language understanding and robust instruction generalization under constrained compute budgets, Motif-2-12.7B builds upon Motif-2.6B with the integration of Grouped Differential Attention (GDA), which improves representational efficiency by disentangling signal and noise-control attention pathways. The model is pre-trained on 5.5 trillion tokens spanning diverse linguistic, mathematical, scientific, and programming domains using a curriculum-driven data scheduler that gradually changes the data composition ratio. The training system leverages the MuonClip optimizer alongside custom high-performance kernels, including fused PolyNorm activations and the Parallel Muon algorithm, yielding significant throughput and memory efficiency gains in large-scale distributed environments. Post-training employs a three-stage supervised fine-tuning pipeline that successively enhances general instruction adherence, compositional understanding, and linguistic precision. Motif-2-12.7B demonstrates competitive performance across diverse benchmarks, showing that thoughtful architectural scaling and optimized training design can rival the capabilities of much larger models.\\
\textbf{Hugging Face}: \url{https://huggingface.co/Motif-Technologies}\\
\textbf{Chat service}: \url{https://chat.motiftech.io/}

\end{abstract}

\section{Introduction}

Recent years have seen rapid progress in large language models (LLMs), with both proprietary and open-weight systems such as GPT-5~\cite{gpt5systemcard}, Claude-4~\cite{Anthropic2025Claude4SC}, Grok-4~\cite{xAI_Grok4_2025}, Gemini-2.5~\cite{comanici2025gemini}, Qwen-3~\cite{yang2025qwen3}, DeepSeek-r1~\cite{guo2025deepseek}, and Kimi K2~\cite{team2025kimi} achieving remarkable results across diverse domains, including mathematics, programming, STEM reasoning, medicine, and law. The scale and breadth of these systems have begun to redefine the boundaries of what language models can represent and reason about, signaling a shift from pattern recognition toward more structured and analytical forms of intelligence. Despite these advances, research on competitive capability under constrained model sizes and limited computational resources remains relatively sparse. Understanding how to push the limits of LLMs' ability within efficiency constraints has emerged as an important challenge for the next generation of language models.

We introduce Motif-2-12.7B(-Base/Instruct), an open-weight foundational language model designed to support instruction following, multi-step reasoning, and general-purpose language understanding.
At its core is Grouped Differential Attention (GDA)~\cite{lim2025grouped}, a new mechanism that reorganizes attention computation to better capture multi-granular token interactions while preserving scalability. 
Motif-2-12.7B achieves competitive results across a broad range of reasoning and instruction-following tasks, showing that architectural innovation and effective training strategies can lead to meaningful improvements even without relying on extremely large parameter sizes. 
With its open availability and transparent training methodology, Motif-2-12.7B would serve as a strong baseline for future research on scalable attention and efficient reasoning.

Compared to its predecessor Motif-2.6B~\citep{lim2025motif}, Motif-2-12.7B represents a significant evolution in scale and training methodology. It is obtained by upscaling Motif-2.6B and pre-training on a substantially larger and more diverse corpus, yielding broader linguistic coverage and deeper knowledge representation across general English, Question Answering, chat, multilingual, scientific, mathematical, and coding domains. Most importantly, we introduce Grouped Differential Attention (GDA), which enhances the model’s capacity to capture complex token dependencies and improves efficiency without additional computational cost. Complemented by strengthened data curation (rigorous filtering, deduplication, novel synthetic data generation, and domain-balanced sampling) and an expanded token budget, Motif-2-12.7B exhibits consistently stronger performance on standard evaluations while preserving continuity with Motif-2.6B.

Pre-training of Motif-2-12.7B-Base was conducted over a corpus total 5.5 trillion tokens, designed to support scalable reasoning and generalization. We employed a linear curriculum scheduler to modulate the data composition throughout training. Early phases consisted predominantly of general-domain English text, with a gradual and controlled increase in the proportion of STEM, mathematical, and programming data. This approach encouraged early convergence on core linguistic fluency while promoting progressive adaptation to more structured reasoning tasks. Optimization was carried out using the MuonClip optimizer~\citep{team2025kimi}, which was chosen for its adaptive gradient scaling properties under large-batch conditions. To fully leverage hardware efficiency, we implemented a set of high-performance parallelism kernels for optimizing MuonClip’s update rule, significantly improves throughput and stability for long-sequence training. This combination of curriculum-aware data scheduling, specialized optimization, and systems-level acceleration enabled us to scale pre-training effectively without compromising convergence quality or compute efficiency.

Following pre-training, Motif-2-12.7B-Instruct further went through a post-training phase consisting of three-stage supervised fine-tuning (SFT) to enhance its general instruction-following and domain-specific abilities. The process began with large-scale alignment on diverse instruction datasets to establish broad conversational competence, then progressively incorporated curated and synthetic data emphasizing compositional reasoning, algorithmic thinking, mathematics, and code generation. In the final refinement stage, selective data pruning was applied to improve diversity and maintain linguistic fluency. Through this gradual and data-driven approach, Motif-2-12.7B-Instruct developed stronger generalization and more coherent reasoning behaviors across a wide range of tasks.

\textbf{Organization of the technical report.}
Section~2 introduces the architectural design of Motif-2-12.7B, detailing its scaling methodology and integration of key components.
Section~3 describes the pretraining data composition, training procedure, and evaluation results of the base model.
Section~4 presents system-level optimizations developed to improve large-scale training efficiency.
Section~5 outlines the three-stage supervised fine-tuning pipeline used to construct the Motif-2-12.7B-Instruct model, along with its corresponding evaluation results.
Finally, Section~6 concludes the report with a summary of findings and future directions.

\section{Architecture}

To expand model capacity while preserving the pretrained representation space, we employed a width expansion strategy that scales internal dimensions without altering the model’s learned function. Motif-2-12.7B was derived from Motif-2.6B through the Scaling Smart Hypercloning procedure~\cite{samragh2024scaling}, in which the model is scaled by an integer multiple through direct weight replication. This method enlarges the network while strictly preserving the original parameter topology and initialization statistics, thereby ensuring functional continuity between generations. Rather than re-initializing or retraining new components from scratch, hypercloning reproduces the pretrained activation tensors exactly, allowing the expanded model to inherit both the structural and representational properties of Motif-2.6B.

To improve representational disentanglement and enhance fine-grained control over information routing within attention, we incorporated the Grouped Differential Attention (GDA) mechanism~\cite{lim2025grouped} during the hypercloning stage. This mechanism enables different subsets of attention heads to specialize in either amplifying salient signals or suppressing residual noise, thereby improving the overall quality and efficiency of learned representations. During this stage, we integrated GDA directly into the cloned architecture, organizing the attention heads into asymmetric groups with a 4:1 ratio between signal-preserving and noise-control groups, respectively. Determining this ratio was a critical design choice, as the GDA mechanism requires balancing the proportion of signal and noise heads for optimal performance. Through preliminary experiments exploring various configurations, we empirically found that a 4:1 ratio provided the best balance—allocating more heads to salient information propagation while maintaining stability and regularization through the smaller noise-control group. This integration represents the first large-scale application of GDA within the Motif architecture, validating its scalability and effectiveness in production-scale training.

Following the width-preserving hypercloning and GDA integration, we expanded the model depth using the LLaMA-Pro framework~\cite{wu2024llama}. This step increased the number of layers in a structurally consistent manner while preserving RMS normalization~\cite{zhang2019root}, Rotary Positional Embedding~\cite{su2024roformer}, and cross-layer residual scaling inherited from Motif-2.6B. The activation PolyNorm~\cite{zhuo2024polynomial} and the tokenizer from Motif-2.6B were retained without modification. The resulting architecture, referred to ``2'' in the middle of its name `Motif-2-12.7B', therefore represents a continuous and functionally extended successor of its predecessor Motif-2.6B. The detailed architectural configurations are summarized in Table~\ref{tab:archi-configs}.

\begin{table}
\centering
\vspace{3pt}
\begin{tabular}{l|c}
    \toprule
    \textbf{Model size} & \textbf{12.7B} \\
    \midrule
    Hidden dimension & 4,096 \\
    Number of layers & 40 \\
    Feed-Forward dimension & 16,384 \\
    Number of heads & 40 \\
    Number of KV heads & 16 \\
    Number of Noise heads & 8 \\
    \midrule
    Attention Mechanism & Grouped Differential Attention \\
    Activation Function & PolyNorm Activation \\
    Max Sequence Length & 32,768 \\
    Positional Embeddings & RoPE ($\theta=1,000,000$) \\
    Vocab size & 219,520 \\
    Tied word embedding & False \\
    \bottomrule
\end{tabular}
\captionof{table}{The detailed hyper-parameters of Motif-2-12.7B.}
\label{tab:archi-configs} 
\end{table}


\section{Pre-training}
This section provides an overview of the pre-training data construction, describes the pre-training methodology, and presents benchmark evaluations of the base models.

\subsection{Pre-training Data}
We constructed a large-scale and high-quality pre-training corpus for Motif-2-12.7B by aggregating a broad spectrum of open-source and in-house datasets spanning diverse domains and modalities. The corpus encompasses general web data (e.g., Nemotron-CC~\cite{su2024nemotron}, Wikipedia, Stack Exchange~\cite{tang2024txt360}, and Facebook Recycling the Web~\cite{nguyen2025recycling}), multilingual and Korean-language resources (Fineweb2~\cite{penedo2025fineweb2pipelinescale}, Translated DiverseQA~\cite{su2024nemotron}, and extended in-house Korean corpus), as well as specialized domains such as reasoning and knowledge-intensive tasks (DiverseQA~\cite{su2024nemotron}, Synthetic QA Dataset), scientific and mathematical content (FineWeb2~\cite{penedo2025fineweb2pipelinescale}, OctoThinker MegaMath~\cite{wang2025octothinker}, Yulan Math~\cite{hu2024yulan}, arXiv), and code-centric corpora (OpenCoder-LLM Dataset~\cite{huang2024opencoder}, Yulan Code~\cite{hu2024yulan}).

The model was trained on 5.5 trillion tokens drawn from this corpus, encompassing general English, question answering, conversational, multilingual, scientific, mathematical, and programming domains. This large-scale training setup provides a balanced and robust foundation for advancing both linguistic and reasoning capabilities.

\subsection {Pre-training Stage and Recipe}
The pre-training of the Motif-2-12.7B series followed three core strategies designed to balance efficiency, diversity, and reasoning capability.

\paragraph{Dataset Mixture Scheduling}
We adopted a progressive dataset mixture scheduling strategy, similar to that used in Motif-2.6B. The early training phase primarily consisted of English web-scale corpora, while reasoning, mathematical, and code-related datasets were introduced in small proportions. Their relative ratios were then gradually increased over the course of training. The dataset mixture was dynamically adjusted at each training step, analogous to how a learning-rate scheduler modulates optimization dynamics. This approach enabled a smooth transition from general linguistic learning to higher-order reasoning without destabilizing early-stage convergence.

\paragraph{Large-Batch Training}
Training was conducted with large effective batch sizes, starting from 16 million tokens and progressively scaling up to 80 million tokens. We employed the Muon-Clip optimizer, originally introduced in the Kimi K2 \cite{team2025kimi}, and further developed an internally optimized variant tailored for parallel and distributed training settings.
This enhanced Muon implementation features improved gradient stability, communication efficiency, and large-batch scalability, which together enable high-throughput optimization under multi-node environments. Details of the parallel and distributed optimization design are provided in a later section.

\paragraph{Reasoning Annealing Stage}
In the final annealing phase, we gradually increased the proportion of reasoning data while capping it below 10\% of total tokens to prevent overfitting and mode collapse toward narrow reasoning distributions. Notably, within this reasoning subset, mathematical data were weighted more heavily than code data, which is a deliberate choice aimed at strengthening symbolic abstraction and quantitative reasoning. Although this allocation is unconventional, we found it conducive to improved general reasoning robustness without degrading linguistic fluency. 
We used a peak learning rate of \( 4 \times 10^{-4} \), which was decayed to approximately 10\% of the value using the Warmup–Stable–Decay scheduler~\cite{hu2024minicpmunveilingpotentialsmall}. The annealing stage was conducted over roughly 1 trillion training tokens, during which the maximum sequence length was gradually increased from 4,096 to 16,384 tokens toward the end of training to enable smooth adaptation to longer-context understanding.

\subsection{Pre-training Evaluation}

\paragraph{Evaluation Setup}

To evaluate the performance of the Motif-2-12.7B-Base, we conducted a comprehensive evaluation across multiple capability areas. We compared the models developed in this study against two leading open-source foundation models, Qwen3 and Gemma3~\cite{team2025gemma}. These two models represent state-of-the-art open weight models similar scale and learning methods. To ensure fair and consistent comparison, we used the officially reported scores from each model’s technical report. All evaluations for Motif-2-12.7B-Base were conducted using greedy decoding.

\noindent\textbf{General Knowledge}
\quad We evaluate benchmarks such as MMLU \cite{hendrycks2009measuring}, MMLU-Redux \cite{gema2025we}, MMLU-Pro \cite{wang2024mmlu}, BBH \cite{suzgun2023challenging}, ARC-Challenge, and ARC-Easy \cite{clark2018think} to measure the model’s breadth of factual knowledge, general reasoning ability. Motif-2-Base achieves solid performance on the MMLU and MMLU-Redux, surpassing Gemma-3 12B and approaching Qwen-3 14B/32B levels. Notably, it records the highest score on MMLU-Pro highlighting strong reasoning capability.

\noindent\textbf{Math \& Scientific Reasoning}
\quad GSM8K \cite{cobbe2021training}, MATH \cite{hendrycks2024measuring}, GPQA, GPQA-Diamond \cite{rein2024gpqa}, and SuperGPQA \cite{du2025supergpqa} are used to assess mathematical reasoning, numerical precision, and scientific understanding requiring multi-step logical inference. Motif-2-Base exhibits clear strength, recording 94.9 on GSM8K and 73.6 on MATH, exceeding all open-weight baselines of comparable scale. It also attains competitive results on GPQA and SuperGPQA, showing superior performance among models of similar scale. 

\noindent\textbf{Code Tasks}
\quad HumanEval \cite{chen2021evaluating}, MBPP \cite{austin2021program}, EvalPlus \cite{liu2023your}, and CRUX-O \cite{gu2024cruxeval} evaluate code generation, reasoning about programs, and functional correctness through test-based execution. Motif-2-Base leads with 65.9 on HumanEval and 81.5 on MBPP, establishing itself as one of the strongest code-capable models among non-proprietary peers.

\noindent\textbf{Common-sense}
\quad HellaSwag \cite{zellers2019hellaswag}, BoolQ \cite{clark2019boolq}, PIQA \cite{bisk2020piqa}, SIQA \cite{sap2019socialiqa}, WinoGrande \cite{sakaguchi2021winogrande}, DROP \cite{dua2019drop}, TriviaQA \cite{joshi2017triviaqa}, and Natural Questions \cite{kwiatkowski2019natural}
are employed to test everyday reasoning, contextual comprehension,
and the model’s ability to infer causal or pragmatic relationships in natural language. Although slightly lower than larger counterparts, the scores remain competitive and do not show significant degradation.

\paragraph{Evaluation Results}

Across diverse benchmarks spanning knowledge, reasoning, coding, and commonsense understanding, Motif-2-12.7B-Base demonstrates strong overall competitiveness with leading open-source foundation models.
Overall, Motif-2-12.7B-Base outperforms both Gemma-3 12B and 27B, as well as Qwen models of comparable size (14B), while approaching the performance of much larger variants such as Qwen-3 32B. These results confirm that Motif-2-12.7B-Base stands as a high-performing and broadly capable foundation model among open-weight models.

\begin{table}[H]
    \centering
    \resizebox{\linewidth}{!}{
    \begin{tabular}{@{}l r r r r r r r r r@{}}
        \toprule
        \textbf{Benchmark} & 
        \textbf{Metric} &
        \multicolumn{1}{c}{\textbf{Motif-2}} &
        \multicolumn{2}{c}{\textbf{Gemma3}} &
        \multicolumn{2}{c}{\textbf{Qwen2.5}} &
        \multicolumn{3}{c}{\textbf{Qwen3}} \\
        & & 12.7B & 12B & 27B & 14B & 32B & 14B & 32B & 30B-A3B \\
        \midrule[0.75pt]
        \textbf{General knowledge} \\
        \midrule[0.75pt]
        MMLU & 5-shot & 78.1 & 74.5 & 78.6 & 79.66 & \underline{83.32} & 81.05 & \textbf{83.61} & 81.38 \\
        MMLU-Redux & 5-shot & 78.68 & - & - & 76.64 & \underline{81.97} & 79.88 & \textbf{83.41} & 81.17 \\
        MMLU-Pro & 5-shot, CoT & \textbf{66.38} & 45.3 & 52.2 & 51.16 & 55.1 & 61.03 & \underline{65.54} & 61.49 \\
        BBH & 3-shot, CoT & 81.34 & 72.6 & 77.7 & 78.18 & \underline{84.48} & 81.07 & \textbf{87.38} & 81.54 \\
        ARC-Easy & 0-shot & 84.1 & \underline{88.3} & \textbf{89.0} & - & - & - & - & - \\
        ARC-Challenge & 25-shot & \underline{69.6} & 68.9 & \textbf{70.6} & - & - & - & - & - \\
        \midrule[0.75pt]
        \textbf{Math \& Scientific Reasoning} \\
        \midrule[0.75pt]
        \multirow{2}{*}{GSM8k} & 4-shot, CoT & \textbf{93.85} & - & - & 90.22 & 92.87 & 92.49 & \underline{93.40} & 91.81 \\
         & 8-shot, CoT & \textbf{94.92} & 71.0 & \underline{82.6} & - & - & - & - & - \\
        MATH & 4-shot, CoT & \textbf{73.62} & 43.3 & 50.0 & 55.64 & 57.7 & \underline{62.02} & 61.62 & 59.04 \\
        GPQA & 5-shot, CoT & 42.18 & - & - & 32.83 & \underline{47.97} & 39.9 & \textbf{49.49} & 43.94 \\
        GPQA-Diamond & 5-shot, CoT & \textbf{42.92} & \underline{25.4} & 24.3 & - & - & - & - & - \\
        SuperGPQA & 5-shot, CoT & 32.68 & - & - & 30.68 & 33.55 & \underline{34.27} & \textbf{39.78} & 35.72 \\
        \midrule[0.75pt]
        \textbf{Coding Tasks} \\
        \midrule[0.75pt]
        HumanEval & 0-shot & \textbf{65.9} & \underline{48.8} & 45.7 & - & - & - & - & - \\
        MBPP & 3-shot & \textbf{81.5} & 60.4 & 65.6 & 69.0 & 73.6 & 73.4 & \underline{78.2} & 74.4 \\
        EvalPlus & 0-shot & \underline{72.22} & - & - & 60.7 & 66.25 & 72.05 & \textbf{72.23} & 71.45 \\
        CRUX-O & 1-shot & 63.1 & - & - & 61.1 & 67.8 & \underline{68.6} & \textbf{72.5} & 67.2 \\
        \midrule[0.75pt]
        \textbf{Common Sense} \\
        \midrule[0.75pt]
        HellaSwag & 10-shot & 84.0 & \underline{84.2} & \textbf{85.6} & - & - & - & - & - \\
        BoolQ & 0-shot & 78.5 & \underline{78.8} & \textbf{82.4} & - & - & - & - & - \\
        PIQA & 0-shot & 81.6 & \underline{81.8} & \textbf{83.3} & - & - & - & - & - \\
        SIQA & 0-shot & \underline{53.8} & 53.4 & \textbf{54.9} & - & - & - & - & - \\
        DROP & 1-shot & 69.9 & \underline{72.2} & \textbf{77.2} & - & - & - & - & - \\
        TriviaQA & 5-shot & 72.2 & \underline{78.2} & \textbf{85.5} & - & - & - & - & - \\
        Natural Questions & 5-shot & 29.6 & \underline{31.4} & \textbf{36.1} & - & - & - & - & - \\
        WinoGrande & 5-shot & \textbf{79.6} & 74.3 & \underline{78.8} & - & - & - & - & - \\
        \midrule[0.75pt]
        \multirow{2}{*}{\textbf{Average}} &  & \textbf{71.53} & 63.87 & \underline{67.96} & - & - & - & - & - \\
         &  & \underline{69.42} & - & - & 62.35 & 67.69 & 67.81 & \textbf{71.54} & 68.10 \\
        \bottomrule
    \end{tabular}
    }
    \caption{Performance comparison across Motif-2, Gemma3, Qwen2.5, and Qwen3 families. Bold indicates the highest score per row; underline indicates the second-highest.}
    \label{tab:benchmark_comparison_gemma3_base}
\end{table}

\section{Training System Optimization}

In this section, we describe how training efficiency was significantly enhanced through the integration of novel kernel fusion and parallel muon techniques.

\subsection{Infrastructure and Precision}

We employed \texttt{SkyPilot}~\cite{yang2023skypilot} as the orchestration framework, enabling seamless multi-node and multi-cluster provisioning, and unified resource abstraction across multiple compute environments. For the core training stack, we adopted \texttt{TorchTitan}~\cite{liang2024torchtitan}, a PyTorch-native distributed training platform that provides modular 3D/4D parallelism (data, tensor, pipeline, and context), advanced optimizations such as FP8 all-gather and activation checkpointing.
Pretraining was performed on 400 H100 gpus for approximately 272K hours, using FP8 precision to maximize computational efficiency. For FP8 training, we applied row-wise scaling and keeps gradients in BF16 precision during the backward pass.

\subsection{Kernel Fusion}
During training, we implemented PolyNorm, a polynomial activation function introduced at Motif-2.6B~\citep{lim2025motif}, as a fused CUDA kernel and further fused it with the subsequent elementwise multiplication in the FeedForward layer. PolyNorm is composed of lightweight sub-operations—elementwise scaling, reduction, and normalization—that are typically memory-bound and hence benefit from kernel fusion, which reduces redundant memory traffic and kernel launch overhead.

\texttt{torch.compile}~\footnote{\url{https://docs.pytorch.org/docs/stable/generated/torch.compile.html}} is a high-level API that enables graph-level optimization and automatic GPU kernel fusion with almost no code modification.
While it provides substantial performance gains, we observed that hand-written fused CUDA kernels can still deliver superior throughput.

\begin{table}[t]
\centering
\small
\resizebox{\textwidth}{!}{%
\begin{tabular}{@{} l cc cc @{}}
\toprule
\multirow{2}{*}{\textbf{Operation}} &
\multicolumn{2}{c}{\textbf{Forward Speedup}} &
\multicolumn{2}{c}{\textbf{Backward Speedup}} \\
\cmidrule(lr){2-3} \cmidrule(lr){4-5}
& \textbf{vs Naive} & \textbf{vs \texttt{torch.compile}} & \textbf{vs Naive} & \textbf{vs \texttt{torch.compile}} \\
\midrule
\makecell[l]{PolyNorm} & 30.29 & 1.53 &	43.90 & 4.77 \\
\makecell[l]{PolyNorm + Elementwise Multiplication} & 23.38 & 1.29 & 27.92 & 3.33 \\
\bottomrule
\end{tabular}
}
\caption{
Performance comparison of fused kernels against baseline implementations on an H200 GPU. 
All experiments use BF16 precision under PyTorch 2.8.
Benchmarks were conducted with hidden sizes of 8K and 16K, sequence lengths from 1K to 8K, and batch sizes between 1 and 4. 
Reported speedups are geometric means across all configurations. 
“Naive” refers to a direct implementation using standard PyTorch APIs.
}
\label{tab:fused_kernel_perf}
\end{table}

Table~\ref{tab:fused_kernel_perf} compares our fused CUDA kernels against both the naive PyTorch implementation and the \texttt{torch.compile}-optimized version. 
While \texttt{torch.compile} already provides substantial acceleration—yielding up to $12.9\times$ speedup over the naive baseline—our hand-written fused kernels achieve even higher throughput in both forward and backward passes. 
This demonstrates that manual fusion can further exploit the remaining optimization headroom, particularly for normalization layers dominated by small, memory-bound operations.
The full implementation of Fused PolyNorm kernels are available at
\url{https://huggingface.co/Motif-Technologies/activation}.

\subsection{Parallel Muon}
\label{sec:parallel-muon}
Liu et al. introduced Distributed Muon~\cite{liu2025muon}, which applies a ZeRO-1–based parameter partitioning scheme to the Muon algorithm. Unlike element-wise optimizers such as AdamW, Muon requires access to the full gradient matrices to perform the Newton–Schulz iterations.
Distributed Muon addresses this by performing an all-gather to reconstruct the full matrix before each Newton–Schulz iteration, computing the update locally, and then discarding all shards except the one that belongs to each rank. While this approach resolves the full-matrix dependency of Muon, it does not parallelize the Newton–Schulz computation itself; as a result, the iteration steps are redundantly executed on all ranks.

There exist several ideas and implementations that aim to overcome the limitations of Distributed Muon. One notable approach is from Essential AI\footnote{\url{https://www.essential.ai/research/infra}},
where they propose employing All-to-All communication to re-shard gradients across ranks, thereby eliminating the redundant computations inherent in Distributed Muon.

We began developing our own implementation of Muon, extending Essential AI’s idea with pipelining and general sharding support (e.g., TP + HSDP), which we refer to as Parallel Muon.
Parallel Muon distributes computational workloads across shard ranks and executes them concurrently.
In addition, it schedules communication and computation operations to fully utilize the accelerators.
While some existing implementations of Muon support FSDP, none provide support for hybrid parallel configurations.

\begin{algorithm}
\caption{Parallel Muon}
\label{alg:parallel_muon}
\setstretch{1.05}
\textbf{Require:} DP sharded gradients $\mathbf{G}$, DP sharded parameters $\mathbf{P}$, rank $\mathbf{r}$
\begin{algorithmic}[1]
\State \texttt{// 1. Divide $\mathbf{G}$ into subsets and assign to each rank}
\State $\mathbf{G}' \gets \text{assign}(\mathbf{G}, \mathbf{r})$
\State \texttt{// 2. All2All communicate to gather $\mathbf{G}'$ across DP shard group}
\State \texttt{// As a result, each rank get unsharded matrices of $\mathbf{G}'$}
\State $\mathbf{G}'_{\text{full}} \gets \text{all2all\_gather}(\text{recv=}\mathbf{G'}, \text{send=}\mathbf{G} \setminus \mathbf{G}')$
\State \texttt{// 3. Calculate Newton-Schulz of unsharded gradients assigned to rank $\mathbf{r}$}
  \State $\mathbf{U}'_{\text{full}} \gets \text{Newton-Schulz}(\mathbf{G}'_{\text{full}})$
\State \texttt{// 4. All2All communicate to scatter $\mathbf{U}'_{\text{full}}$ across DP shard group}
\State \texttt{// As a result, earch rank get sharded matrices of $\mathbf{U}'_{\text{full}}$}
\State $\mathbf{U} \gets \text{all2all\_scatter}(\text{recv=}\mathbf{U}'_{\text{other}}, \text{send=}\mathbf{U}'_{\text{full}})$
\State \texttt{// 5. Apply DP sharded $\mathbf{U}$ to $\mathbf{P}$}
\State $\mathbf{P'} \gets \text{apply\_update}(\mathbf{P}, \mathbf{U})$
\State \textbf{return $\mathbf{P'}$}
\end{algorithmic}
\end{algorithm}

Algorithm~\ref{alg:parallel_muon} describes the parallel version of Muon under Fully Sharded Data-Parallel (FSDP) training. Each rank first takes a subset of gradients and then participates in an All2All communication to gather the full unsharded gradients assigned to it. The Newton–Schulz iteration is then applied independently on each gathered subset to compute the update matrices. The resulting unsharded updates are scattered back to the corresponding ranks via another All2All operation, producing sharded update tensors. Finally, each rank applies its local update to the corresponding sharded parameters. This procedure enables concurrent computation across DP-shard ranks without additional memory overhead.

\paragraph*{All-to-All Gather and Scatter}

\begin{figure}
    \centering
    \includegraphics[width=1\linewidth]{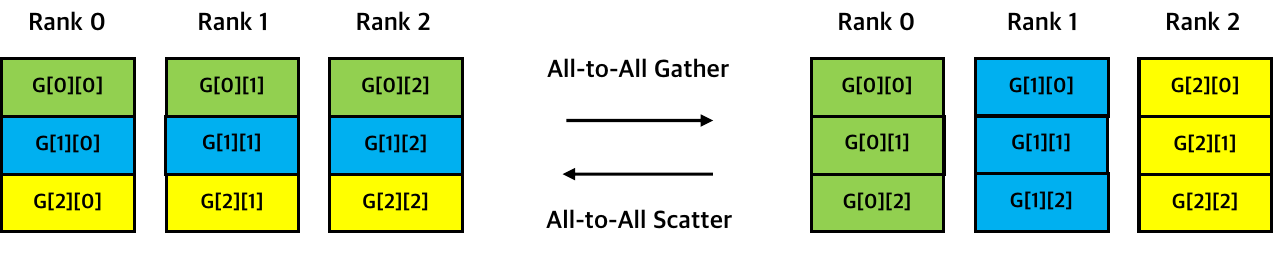}
    \caption{Illustration of the All-to-All gather and scatter process in Parallel Muon. Each rank exchanges its sharded gradients with all other ranks during the gather phase and redistributes the computed results in the scatter phase.}
    \label{fig:all_to_all}
\end{figure}

As discussed earlier in Essential AI’s idea, Parallel Muon also employs All-to-All communication to redistribute both sharded and unsharded gradients across ranks.
The All-to-All gather operation collects the sharded gradients so that each rank obtains its assigned unsharded gradients, while the All-to-All scatter operation redistributes the computed results back to all ranks, as illustrated in Figure~\ref{fig:all_to_all}.

\paragraph*{Pipelined Execution}

\begin{figure}
    \centering
    \includegraphics[width=1\linewidth]{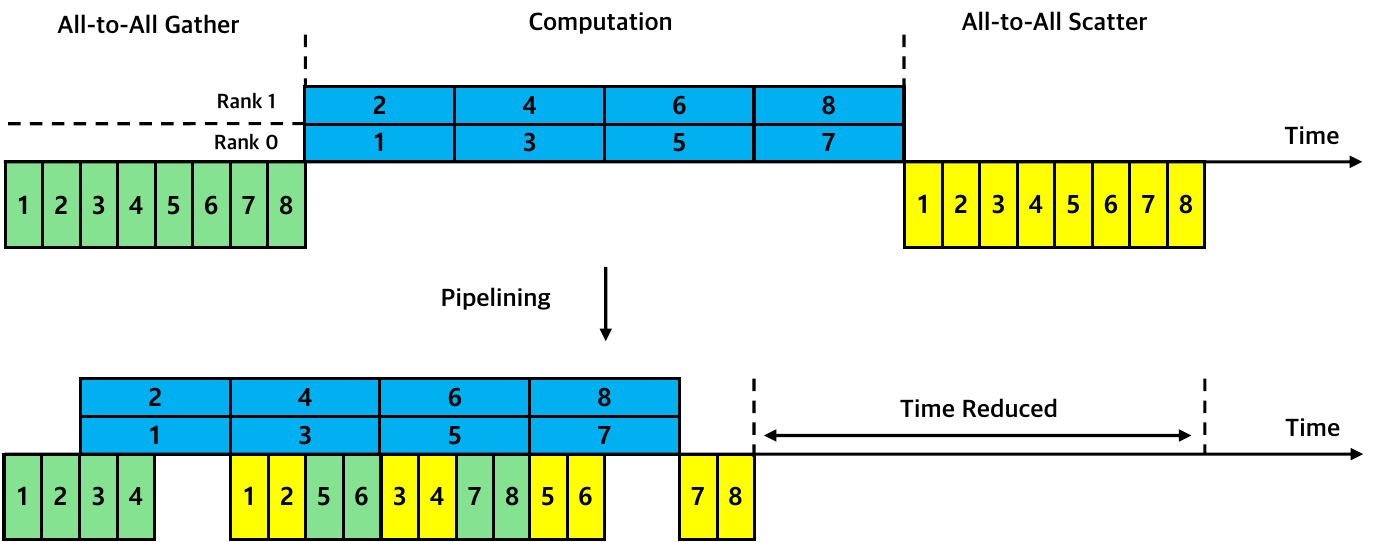}
    \caption{Illustration of pipelined execution of all-to-all gather, computation, and scatter phases. A total of eight gradients are distributed across two ranks and partitioned into chunks of size two. Computation and communication are scheduled in a pipelined manner, allowing overlap between them and improving overall efficiency.}
    \label{fig:pipelining_muon}
\end{figure}

Although Algorithm \ref{alg:parallel_muon} is sound on its own, pipelining its execution enables better performance through communication–computation overlap. As illustrated in Figure~\ref{fig:pipelining_muon}, we group the parameters into chunks and schedule their all-to-all \emph{gather}, \emph{computation}, and \emph{scatter} phases in a pipelined fashion.
Specifically, we first issue the \emph{gather} and \emph{computation} operations for a few initial chunks to warm up the pipeline. Then, for each subsequent chunk, we overlap \emph{scatter} of the previous results, \emph{gather} of the next inputs, and the \emph{computation} of the current chunk.
This scheduling can hide communication latency by maintaining a steady flow of computation and communication, thereby improving throughput and reducing idle time.

In addition to performance improvements, pipelining also helps reduce peak memory usage. 
Since the full output matrices produced by each computation can only be deallocated after their corresponding \emph{scatter} phase completes, a non-pipelined execution must keep multiple full matrices in memory simultaneously. 
By overlapping computation and scatter, the pipelined schedule allows completed full matrices to be released earlier, thus lowering the overall memory footprint.

\paragraph{Chunk Size Trade-off}
While pipelining improves both performance and memory efficiency, one important consideration is the choice of \emph{chunk size}. 
A smaller chunk size allows finer-grained overlap between computation and communication, which tends to improve performance and reduce memory usage. 
However, all-to-all communication becomes less efficient when messages are too small, making larger chunks more favorable for bandwidth utilization. 
In practice, the benefit from improved bandwidth utilization with larger chunks often outweighs the performance gain from finer overlap, resulting in an optimal chunk size that balances both effects. 
We empirically find that a chunk size of 32 achieves the best overall trade-off between performance and memory usage.

\paragraph*{Mitigating Rank Imbalance}
To mitigate workload imbalance among ranks, we balanced the total computational load assigned to each rank.
Because the all-to-all communication in gather and scatter stages requires synchronization among all participating ranks, any imbalance in computation time would lead to idle waiting and reduced throughput.
We therefore sorted all gradients according to the number of floating-point operations (FLOPs) required for their respective Newton–Schulz computations and assigned them to ranks in a round-robin fashion.
This simple yet effective heuristic ensures that each rank receives a roughly equal total workload, reducing idle time and improving parallel efficiency.
While this strategy does not guarantee perfectly balanced execution, it provides a good practical trade-off between load balance and scheduling overhead.

\paragraph*{Support for Hybrid Parallel Configurations}
To support various strategies beyond FSDP, such as TP or TP + HSDP, Parallel Muon dynamically inspects the placement of gradient tensors at runtime, constructs corresponding sub–process groups for gather and scatter operations, and computes the shard mapping that defines which tensor slice is held by each rank.

Since Parallel Muon preserves the semantics of the original Newton–Schulz iterations, it directly leverages existing kernel optimizations, including Flash Muon~\cite{lin2025flash}. The full implementation of Parallel Muon is available at \url{https://huggingface.co/Motif-Technologies/optimizer}.

\paragraph*{Parallel Muon Evaluation}
We evaluate the performance of Parallel Muon under various configurations and compare it against the Distributed Muon baseline. 
All experiments are conducted on a single node equipped with eight NVIDIA H200 GPUs, using Fully Sharded Data Parallel (FSDP) with eight ranks. 
The model used for evaluation is Motif-2-12.7B, and all communication and GEMM operations are performed in BF16 precision.

We consider the following configurations:
\begin{enumerate}
    \item \textbf{Distributed Muon} – the baseline implementation, extended with Flash Muon\cite{lin2025flash} for fair comparison with Parallel Muon. 
    Distributed Muon reconstructs the full parameter matrix via all-gather and redundantly executes the Newton–Schulz iterations on all ranks.
    \item \textbf{Parallel Muon (non-pipelined)} – our implementation of Parallel Muon without pipelining, where computation and communication are executed sequentially.
    \item \textbf{Parallel Muon (pipelined)} – Parallel Muon with pipelined gather–compute–scatter scheduling to enable communication–computation overlap.
    \item \textbf{Parallel Muon (pipelined + FLOPs-sorted)} – the full version, which additionally applies FLOPs-based gradient sorting and round-robin scheduling to balance computation across ranks.
\end{enumerate}
We measure throughput, step time, and peak memory usage for each configuration to quantify the impact of pipelining and load balancing. 

Table~\ref{tab:parallel_muon_results} summarizes the performance of Distributed Muon and the different Parallel Muon configurations.

First, simply adopting the non-pipelined Parallel Muon configuration (row 2) achieves a substantial performance of 571 Tera floating operations per second per GPU (TFLOPS/GPU), a 7.1$\times$ throughput improvement over the baseline Distributed Muon (80 TFLOPS/GPU). This significant gain is attributed to distributing the computational workload across the 8 GPUs.

Interestingly, introducing pipelining (row 3) initially results in a performance degradation, with throughput dropping by approximately 15.8\% compared to the non-pipelined version. This is caused by workload imbalance, as the per-chunk all-to-all communication (for the small, size 32 chunks) acts as a synchronization barrier across ranks. This is strongly supported by the results from the fourth configuration: applying FLOPs-based parameter sorting (row 4) rectifies this imbalance, boosting performance significantly by over 21\%.

It is also important to note that the throughput of the fully optimized (pipelined + sorted) version is only marginally higher—by about 2.1\%—than that of the non-pipelined version. This relates to the chunk size trade-off; the non-pipelined approach effectively uses a single, large chunk (equal to the total number of parameters), thus benefiting from high bandwidth utilization on its single communication step while also avoiding the repeated synchronization overheads inherent to fine-grained pipelining.

Despite this small throughput difference, the primary benefit of the pipelined approach is its drastically lower memory footprint. The non-pipelined version's peak memory usage is over 4.1 times that of the pipelined (non-sorted) configuration and over 3 times that of the fully optimized version, making pipelining a crucial optimization for memory efficiency.

Another open-sourced Muon implementation example\footnote{\url{https://github.com/microsoft/dion/blob/main/dion/muon.py}} is provided in Dion~\cite{ahn2025diondistributedorthonormalizedupdates}. This work is also inspired by Essential AI’s approach, and its key idea is to avoid the full-matrix reconstruction required by Muon. However, this implementation does not support hybrid configurations such as FSDP2 combined with tensor parallelism (TP).

\begin{table}[H]
\centering
\small
\caption{Performance comparison of Distributed Muon and Parallel Muon configurations on 8$\times$H200 GPUs. All experiments use BF16 precision under FSDP.}
\label{tab:parallel_muon_results}
\resizebox{\textwidth}{!}{%
\begin{tabular}{@{} l c c c c r r r @{}}
\toprule
\textbf{Configuration} & \textbf{Pipelining} & \textbf{Sort Param.} & \textbf{Chunk Size} & \textbf{Time (ms)} & \textbf{Peak Mem (MB)} & \textbf{TFLOPS per GPU} \\
\midrule
\makecell[l]{Distributed Muon} 
  & -- & -- & -- & 1574.67 & 832   & 80  \\
\makecell[l]{Parallel Muon} 
  & X  & X  & -- & 221.27  & 11904 & 571 \\
\makecell[l]{Parallel Muon} 
  & O  & X  & 32  & 262.83  & 2894  & 481 \\
\makecell[l]{Parallel Muon} 
  & O  & O  & 32  & 216.46  & 3904  & 583 \\
\bottomrule
\end{tabular}
}
\end{table}

\section{Post-training}

\subsection{Supervised Fine-tuning}
Following the completion of pre-training, Motif-2-12.7B-Instruct went through a \textbf{three-stage supervised fine-tuning (SFT)} process designed to enhance its instruction-following, reasoning, and domain specialization capabilities. 
Unlike Motif-2.6B, which primarily focused on general instruction adherence, each stage explicitly targeted multi-step reasoning, code synthesis, and mathematical problem-solving through carefully curated and synthesized datasets. 
The three-stage SFT pipeline progressively refines the model’s capabilities—from large-scale general instruction learning (Stage 1), to reasoning- and domain-oriented enhancement through synthetic data (Stage 2), and finally to data-pruned targeted refinement (Stage 3).

\subsubsection{Stage 1: Large-Scale Supervised Fine-tuning}

The first fine-tuning stage focused on establishing strong general instruction-following performance through large-scale training. 
We trained on approximately 28 million samples, drawn from both open-source and proprietary datasets, with domain ratios adjusted to maintain a balanced representation across categories. 
Training was conducted with sequence packing up to 16,384 tokens to enable efficient utilization of long-context information. 
All parameters were unfrozen during this stage, allowing full adaptation of model representations to downstream supervision.

\paragraph*{Training Setup}
Training was performed using the Muon-Clip optimizer with a base learning rate of $2\times10^{-5}$, weight decay of 0.1, and a global batch size of 32 million tokens. 
A cosine learning rate scheduler was employed with a warm-up phase corresponding to 5\% of the total training steps, followed by smooth decay toward zero. 
Sequence packing was enabled up to a maximum sequence length of 16,384 tokens to ensure efficient utilization of GPU memory and to improve long-context training stability. 
All training was conducted in FP8 precision under Fully Sharded Data Parallel (FSDP) with the Parallel Muon optimization kernels as described in Section~\ref{sec:parallel-muon}, thereby enabling stable and high-throughput large-batch training.

\subsubsection{Stage 2: Synthetic and Targeted Fine-tuning}

To further enhance higher-order reasoning and domain-specific capabilities, the second fine-tuning phase employed a mixture of open and synthetic instruction datasets. 
We first consolidated a diverse mixture of high-quality open instruction datasets spanning general conversation, STEM, mathematics, and code generation. 
Each dataset was evaluated based on reasoning density, response completeness, and structural consistency. 
The resulting dataset was designed to maintain domain balance while providing comprehensive coverage of general instruction-following and structured reasoning tasks.

Building upon this foundation, we introduced additional synthetic data generated from both internal and external models to strengthen compositional reasoning, algorithmic problem-solving, and multi-step instruction-following performance. 
The synthetic datasets consisted of:
\begin{itemize}
    \item \textbf{Compositional Reasoning:} Synthetic examples designed to enhance multi-step logical reasoning and structured task decomposition.
    \item \textbf{Algorithmic and Mathematical Skills:} Data emphasizing algorithmic reasoning, code generation, and quantitative problem-solving.
    \item \textbf{Quality-Centric Selection:} High-quality samples curated through multi-stage filtering and automatic quality scoring.
\end{itemize}

This stage adopted a lower learning rate and shorter training schedule relative to Stage 1, functioning as a form of curriculum continuation. 
Through the combination of large-scale instruction data and carefully designed synthetic augmentation, Motif-2-12.7B-Instruct achieved significant improvements in reasoning quality while preserving conversational fluency.

\subsubsection{Stage 3: Data-Pruned Refinement}

In the final stage, we conducted a focused fine-tuning process by selectively excluding a subset of the Stage 2 synthetic data identified as redundant or low-utility. 
This data-pruned refinement stage aimed to reinforce domain diversity and reduce potential overfitting to synthetic distributions by re-training on a carefully filtered subset of Stage 2 data. 
The pruning criteria emphasized diversity, reasoning coherence, and linguistic fidelity, resulting in a leaner but higher-quality dataset. 
This stage served as a lightweight continuation of Stage 2, stabilizing the model’s reasoning and alignment behavior without altering its general conversational fluency.

\subsection{Post-training Evaluation}

\paragraph{Evaluation Setup}
Following the same evaluation protocol as Motif-2-12.7B-Base, we conducted a comprehensive assessment of Motif-2-12.7B-Instruct to evaluate its generalization, reasoning, and instruction-following capabilities. To ensure comparability and fairness, we benchmarked our model against strong open-weight baselines, including the Qwen3 and Gemma3 series, using their officially reported scores from the respective technical reports. All evaluations for Motif-2-12.7B-Instruct were performed with a sampling temperature of 0.6 and a maximum sequence length of 32,768 tokens.

\paragraph{Benchmark Composition}
Our evaluation suite spans a wide range of domains to measure both breadth and depth of model ability. The general task benchmarks include MMLU~\cite{hendrycks2009measuring}, MMLU-Redux~\cite{gema2025we}, GPQA-Diamond~\cite{rein2024gpqa}, LiveBench~\cite{white2024livebench}, and IFEval~\cite{ifeval2024}, which collectively test factual knowledge, reasoning, and alignment quality across diverse subjects.
The mathematics and text reasoning category comprises MATH, MATH-500~\cite{lightman2023let}, AIME24, AIME25~\cite{AIME25}, ZebraLogic~\cite{lin2025zebralogic}, and BBH~\cite{suzgun2023challenging}, emphasizing symbolic reasoning, structured problem-solving, and compositional understanding.
Finally, agentic and programming-oriented benchmarks, including BFCL~v3~\cite{xie2023doremi}, LiveCodeBench~\cite{jain2024livecodebench}, MBPP~\cite{austin2021program}, and HumanEval~\cite{chen2021evaluating}, assess multi-step reasoning, code synthesis, and executional accuracy in practical scenarios.
This diverse set of tasks allows us to holistically evaluate the instruction-following ability and analytical robustness of Motif-2-12.7B-Instruct across both linguistic and technical domains.

\begin{table}[b]
    \centering
    \resizebox{\linewidth}{!}{
    \begin{tabular}{@{}l r r r r r r r r r r r @{}}
        \toprule
        \textbf{Benchmark} & 
        \textbf{Metric} &
        \multicolumn{1}{c}{\textbf{Motif-2}} &
        \multicolumn{1}{c}{\textbf{Qwen2.5}} &
        \multicolumn{6}{c}{\textbf{Qwen3}} \\
        \cmidrule(lr){3-3} \cmidrule(lr){4-4} \cmidrule(lr){5-10} 
        & & 12.7B & 72B & 14B & 14B & 32B & 32B & 30B-A3B & 30B-A3B \\
        & & Instruct & Non-Think & Non-Think & Think & Non-Think & Think & Non-Think & Think  \\
        \midrule[0.75pt]
        MMLU-Redux & 0-shot & 90.02 & 86.8 & 82 & 88.6 & 85.7 & 90.9 & 84.1 & 89.5 \\
        GPQA-Diamond & 0-shot, CoT & 63.6 & 49 & 54.8 & 64 & 54.6 & 68.4 & 54.8 & 65.8 \\
        \makecell[l]{LiveBench \\ 2024-11-25} & - & 33.8 & 51.4 & 59.6 & 71.3 & 59.8 & 74.9 & 59.4 & 74.3 \\
        MATH-500 & 0-shot & 96.8 & 83.6 & 90 & 96.8 & 88.6 & 97.2 & 89.8 & 98 \\
        AIME24 & 0-shot & 72.3 & 18.9 & 31.7 & 79.3 & 31 & 81.4 & 32.8 & 80.4 \\
        AIME25 & 0-shot & 63.6 & 15 & 23.3 & 70.4 & 20.2 & 72.9 & 21.6 & 70.9  \\
        ZebraLogic & - & 69.5 & 26.6 & 33 & 88.5 & 29.2 & 88.8 & 33.2 & 89.5 \\
        BFCLv3 & - & 55.34 & 63.4 & 61.5 & 70.4 & 63 & 70.3 & 58.6 & 69.1 \\
        \makecell[l]{LiveCodeBench v5 \\ (2024.10 - 2025.2)}  & 0-shot & 50.03 & 30.7 & 29 & 63.5 & 31.3 & 65.7 & 29.8 & 62.6\\
        IFEval & strict prompt & 75.78 & 84.1 &  84.8 & 85.4 & 83.2 & 85 & 83.7 & 86.5 \\ 
        \midrule[0.75pt]
        \textbf{Average} & & 67.08 & 50.95 & 54.97 & 77.82 & 54.66 & 79.55 & 54.78 & 78.66 \\
        \bottomrule
    \end{tabular}
    }
    \caption{Performance comparison across Motif-2, and Qwen families.}
    \label{tab:benchmark_comparison_sft_qwen}
\end{table}

\begin{table}[t]
    \centering

    \begin{tabular}{@{}l r r r r@{}}
        \toprule
        \textbf{Benchmark} & 
        \textbf{Metric} &
        \multicolumn{1}{c}{\textbf{Motif-2}} &
        \multicolumn{2}{c}{\textbf{Gemma3}} \\
        \cmidrule(lr){3-3} \cmidrule(lr){4-5}
        & & 12.7B & 12B  & 27B\\
        \midrule[0.75pt]

        MMLU & 0-shot & 86.11 & 71.9 & 76.9 \\
        BBH & 0-shot & 85.78 & 85.7 & 87.6 \\
        GSM8k & 0-shot, CoT & 96.13& 94.4 & 95.9 \\
        MATH & 0-shot & 97 & 83.8 & 89 \\
        MBPP & 3-shot & 91 & 73 & 74.4 \\
        IFEval & soft prompt & 76.52 & 88.9 & 90.4 \\
        LiveCodeBench v5 & 0-shot & 61.66 & 32 & 39 \\
        HumanEval & 0-shot & 93.2 & 85.4 & 87.8 \\
        \midrule[0.75pt]
        \textbf{Average} & & 83.44 & 72.89 & 75.93 \\
        \bottomrule
    \end{tabular}
    \caption{Performance comparison across Motif-2, Gemma3.}
    \label{tab:benchmark_comparison_sft_gemma}
\end{table}

\paragraph{Evaluation Results}
Detailed quantitative results are summarized in Table~\ref{tab:benchmark_comparison_sft_qwen} and Table~\ref{tab:benchmark_comparison_sft_gemma}. Overall, Motif-2-12.7B-Instruct achieves performance comparable to or exceeding that of similarly sized or larger open-weight models, despite being trained on a substantially smaller dataset. Notably, without relying on computationally heavy and reasoning targeted reinforcement learning, the model demonstrates strong results on high-difficulty reasoning and code-oriented benchmarks such as MATH-500, AIME25, and LiveCodeBench, underscoring the effectiveness of its three-stage SFT curriculum.
While Qwen3 and Gemma3 models benefit from significantly larger training corpora, with Qwen3 trained on 36T tokens and Gemma3 on 12 to 14T tokens, our model attains similar or higher accuracy across several benchmarks using only 5.5T tokens, highlighting the data efficiency and architectural effectiveness of our approach. The integration of Grouped Differential Attention (GDA), curriculum-aware data scheduling, and high-throughput optimization via Parallel Muon collectively contribute to this efficiency, enabling more capable instruction models at a fraction of the compute cost.
These results illustrate that careful architectural design, optimization-aware training infrastructure, and balanced fine-tuning strategies can bridge much of the performance gap traditionally attributed to massive scale, reaffirming the viability of efficient frontier–oriented model design as a path toward high-quality open-weight systems.

\section{Conclusion}

Motif-2-12.7B demonstrates that carefully targeted architectural changes and systems-aware training can deliver strong performance without resorting to extreme parameter counts. By scaling Motif-2.6B with width-preserving hypercloning and Llama Pro depth scaling, integrating Grouped Differential Attention for finer-grained information routing, and coupling curriculum-aware pre-training with a three-stage SFT pipeline, we obtain a model that is both efficient and broadly capable. Our fused PolyNorm kernels and Parallel Muon further show how bespoke systems work unlocks practical throughput and memory gains at long context, making large-scale training more accessible. Looking ahead, we will release Motif-2-12.7B-Reasoning, a reinforcement-learning–enhanced variant that explicitly optimizes multi-step reasoning quality (with a focus on mathematics and code) while maintaining conversational fluency. We hope these open weights, transparent recipes, and forthcoming RL results provide a solid baseline for the community to study scalable attention, efficient training, and principled pathways to stronger reasoning.

\bibliographystyle{plain}

\bibliography{reference}

@article{lim2025grouped,
  title={Grouped Differential Attention},
  author={Lim, Junghwan and Lee, Sungmin and Kim, Dongseok and Cheung, Wai Ting and Kim, Beomgyu and Kim, Taehwan and Lee, Haesol and Lee, Junhyeok and Oh, Dongpin and Park, Eunhwan},
  journal={arXiv preprint arXiv:2510.06949},
  year={2025}
}

@article{lim2025motif,
  title={Motif 2.6 B Technical Report},
  author={Lim, Junghwan and Lee, Sungmin and Kim, Dongseok and Park, Eunhwan and Park, Hyunbyung and Lee, Junhyeok and Cheung, Wai Ting and Choi, Dahye and Her, Jaeheui and Huh, Jaeyeon and others},
  journal={arXiv preprint arXiv:2508.09148},
  year={2025}
}

@article{team2025kimi,
  title={Kimi k2: Open agentic intelligence},
  author={Team, Kimi and Bai, Yifan and Bao, Yiping and Chen, Guanduo and Chen, Jiahao and Chen, Ningxin and Chen, Ruijue and Chen, Yanru and Chen, Yuankun and Chen, Yutian and others},
  journal={arXiv preprint arXiv:2507.20534},
  year={2025}
}

@article{liu2025muon,
  title={Muon is scalable for LLM training},
  author={Liu, Jingyuan and Su, Jianlin and Yao, Xingcheng and Jiang, Zhejun and Lai, Guokun and Du, Yulun and Qin, Yidao and Xu, Weixin and Lu, Enzhe and Yan, Junjie and others},
  journal={arXiv preprint arXiv:2502.16982},
  year={2025}
}

@article{samragh2024scaling,
  title={Scaling smart: Accelerating large language model pre-training with small model initialization},
  author={Samragh, Mohammad and Mirzadeh, Iman and Vahid, Keivan Alizadeh and Faghri, Fartash and Cho, Minsik and Nabi, Moin and Naik, Devang and Farajtabar, Mehrdad},
  journal={arXiv preprint arXiv:2409.12903},
  year={2024}
}

@article{wu2024llama,
  title={Llama pro: Progressive llama with block expansion},
  author={Wu, Chengyue and Gan, Yukang and Ge, Yixiao and Lu, Zeyu and Wang, Jiahao and Feng, Ye and Shan, Ying and Luo, Ping},
  journal={arXiv preprint arXiv:2401.02415},
  year={2024}
}

@misc{lin2025flash,
  author       = {Tianyang Lin},
  title        = {Flash-Muon: An Efficient Implementation of Muon Optimizer},
  year         = {2025},
  url          = {https://github.com/nil0x9/flash-muon}
}

@article{su2024nemotron,
  title={Nemotron-CC: Transforming Common Crawl into a refined long-horizon pretraining dataset},
  author={Su, Dan and Kong, Kezhi and Lin, Ying and Jennings, Joseph and Norick, Brandon and Kliegl, Markus and Patwary, Mostofa and Shoeybi, Mohammad and Catanzaro, Bryan},
  journal={arXiv preprint arXiv:2412.02595},
  year={2024}
}

@misc{tang2024txt360,
  title={Txt360: A top-quality llm pre-training dataset requires the perfect blend},
  author={Tang, Liping and Ranjan, Nikhil and Pangarkar, Omkar and Liang, Xuezhi and Wang, Zhen and An, Li and Rao, Bhaskar and Jin, Linghao and Wang, Huijuan and Cheng, Zhoujun and others},
  year={2024}
}

@article{nguyen2025recycling,
  title={Recycling the Web: A Method to Enhance Pre-training Data Quality and Quantity for Language Models},
  author={Nguyen, Thao and Li, Yang and Golovneva, Olga and Zettlemoyer, Luke and Oh, Sewoong and Schmidt, Ludwig and Li, Xian},
  journal={arXiv preprint arXiv:2506.04689},
  year={2025}
}

@misc{penedo2025fineweb2pipelinescale,
  title={FineWeb2: One Pipeline to Scale Them All -- Adapting Pre-Training Data Processing to Every Language}, 
  author={Guilherme Penedo and Hynek Kydlíček and Vinko Sabolčec and Bettina Messmer and Negar Foroutan and Amir Hossein Kargaran and Colin Raffel and Martin Jaggi and Leandro Von Werra and Thomas Wolf},
  year={2025},
  eprint={2506.20920},
  archivePrefix={arXiv},
  primaryClass={cs.CL},
  url={https://arxiv.org/abs/2506.20920}, 
}

@article{wang2025octothinker,
  title={Octothinker: Mid-training incentivizes reinforcement learning scaling},
  author={Wang, Zengzhi and Zhou, Fan and Li, Xuefeng and Liu, Pengfei},
  journal={arXiv preprint arXiv:2506.20512},
  year={2025}
}

@article{hu2024yulan,
  title={Yulan-mini: An open data-efficient language model},
  author={Hu, Yiwen and Song, Huatong and Deng, Jia and Wang, Jiapeng and Chen, Jie and Zhou, Kun and Zhu, Yutao and Jiang, Jinhao and Dong, Zican and Zhao, Wayne Xin and others},
  journal={arXiv preprint arXiv:2412.17743},
  year={2024}
}

@article{huang2024opencoder,
  title={Opencoder: The open cookbook for top-tier code large language models},
  author={Huang, Siming and Cheng, Tianhao and Liu, Jason Klein and Hao, Jiaran and Song, Liuyihan and Xu, Yang and Yang, J and Liu, Jiaheng and Zhang, Chenchen and Chai, Linzheng and others},
  journal={arXiv preprint arXiv:2411.04905},
  year={2024}
}

@misc{gpt5systemcard,
  author = {{OpenAI}},
  title = {{GPT-5 System Card}},
  howpublished = {\url{https://openai.com/index/gpt-5-system-card/}},
  year = {2025}
}

@misc{Anthropic2025Claude4SC,
  author = {Anthropic},
  title = {{System Card: Claude Opus 4 \& Claude Sonnet 4}},
  howpublished = {\url{https://www-cdn.anthropic.com/4263b940cabb546aa0e3283f35b686f4f3b2ff47.pdf}},
  year = {2025}
}

@techreport{xAI_Grok4_2025,
    author = {{xAI}},
    title = {{Grok 4 Model Card}},
    institution = {{xAI}},
    year = {2025},
    month = {aug},
    note = {{Last updated August 20, 2025}},
    url = {https://data.x.ai/2025-08-20-grok-4-model-card.pdf},
}

@article{comanici2025gemini,
  title={Gemini 2.5: Pushing the frontier with advanced reasoning, multimodality, long context, and next generation agentic capabilities},
  author={Comanici, Gheorghe and Bieber, Eric and Schaekermann, Mike and Pasupat, Ice and Sachdeva, Noveen and Dhillon, Inderjit and Blistein, Marcel and Ram, Ori and Zhang, Dan and Rosen, Evan and others},
  journal={arXiv preprint arXiv:2507.06261},
  year={2025}
}

@article{guo2025deepseek,
  title={Deepseek-r1: Incentivizing reasoning capability in llms via reinforcement learning},
  author={Guo, Daya and Yang, Dejian and Zhang, Haowei and Song, Junxiao and Zhang, Ruoyu and Xu, Runxin and Zhu, Qihao and Ma, Shirong and Wang, Peiyi and Bi, Xiao and others},
  journal={arXiv preprint arXiv:2501.12948},
  year={2025}
}

@article{yang2025qwen3,
  title={Qwen3 technical report},
  author={Yang, An and Li, Anfeng and Yang, Baosong and Zhang, Beichen and Hui, Binyuan and Zheng, Bo and Yu, Bowen and Gao, Chang and Huang, Chengen and Lv, Chenxu and others},
  journal={arXiv preprint arXiv:2505.09388},
  year={2025}
}

@misc{ahn2025diondistributedorthonormalizedupdates,
      title={Dion: Distributed Orthonormalized Updates}, 
      author={Kwangjun Ahn and Byron Xu and Natalie Abreu and Ying Fan and Gagik Magakyan and Pratyusha Sharma and Zheng Zhan and John Langford},
      year={2025},
      eprint={2504.05295},
      archivePrefix={arXiv},
      primaryClass={cs.LG},
      url={https://arxiv.org/abs/2504.05295}, 
}

@misc{hu2024minicpmunveilingpotentialsmall,
      title={MiniCPM: Unveiling the Potential of Small Language Models with Scalable Training Strategies}, 
      author={Shengding Hu and Yuge Tu and Xu Han and Chaoqun He and Ganqu Cui and Xiang Long and Zhi Zheng and Yewei Fang and Yuxiang Huang and Weilin Zhao and Xinrong Zhang and Zheng Leng Thai and Kaihuo Zhang and Chongyi Wang and Yuan Yao and Chenyang Zhao and Jie Zhou and Jie Cai and Zhongwu Zhai and Ning Ding and Chao Jia and Guoyang Zeng and Dahai Li and Zhiyuan Liu and Maosong Sun},
      year={2024},
      eprint={2404.06395},
      archivePrefix={arXiv},
      primaryClass={cs.CL},
      url={https://arxiv.org/abs/2404.06395}, 
}

@article{team2025gemma,
  title={Gemma 3 technical report},
  author={Team, Gemma and Kamath, Aishwarya and Ferret, Johan and Pathak, Shreya and Vieillard, Nino and Merhej, Ramona and Perrin, Sarah and Matejovicova, Tatiana and Ram{\'e}, Alexandre and Rivi{\`e}re, Morgane and others},
  journal={arXiv preprint arXiv:2503.19786},
  year={2025}
}

@article{liang2024torchtitan,
  title={TorchTitan: One-stop PyTorch native solution for production ready LLM pre-training},
  author={Liang, Wanchao and Liu, Tianyu and Wright, Less and Constable, Will and Gu, Andrew and Huang, Chien-Chin and Zhang, Iris and Feng, Wei and Huang, Howard and Wang, Junjie and others},
  journal={arXiv preprint arXiv:2410.06511},
  year={2024}
}

@inproceedings{yang2023skypilot,
  title={$\{$SkyPilot$\}$: An intercloud broker for sky computing},
  author={Yang, Zongheng and Wu, Zhanghao and Luo, Michael and Chiang, Wei-Lin and Bhardwaj, Romil and Kwon, Woosuk and Zhuang, Siyuan and Luan, Frank Sifei and Mittal, Gautam and Shenker, Scott and others},
  booktitle={20th USENIX Symposium on Networked Systems Design and Implementation (NSDI 23)},
  pages={437--455},
  year={2023}
}

@article{zhang2019root,
  title={Root mean square layer normalization},
  author={Zhang, Biao and Sennrich, Rico},
  journal={Advances in neural information processing systems},
  volume={32},
  year={2019}
}

@article{su2024roformer,
  title={Roformer: Enhanced transformer with rotary position embedding},
  author={Su, Jianlin and Ahmed, Murtadha and Lu, Yu and Pan, Shengfeng and Bo, Wen and Liu, Yunfeng},
  journal={Neurocomputing},
  volume={568},
  pages={127063},
  year={2024},
  publisher={Elsevier}
}

@article{zhuo2024polynomial,
  title={Polynomial Composition Activations: Unleashing the Dynamics of Large Language Models},
  author={Zhuo, Zhijian and Wang, Ya and Zeng, Yutao and Li, Xiaoqing and Zhou, Xun and Ma, Jinwen},
  journal={arXiv preprint arXiv:2411.03884},
  year={2024}
}

@article{hendrycks2009measuring,
      title={Measuring massive multitask language understanding, 2021},
      author={Hendrycks, Dan and Burns, Collin and Basart, Steven and Zou, Andy and Mazeika, Mantas and Song, Dawn and Steinhardt, Jacob},
      journal={URL https://arxiv. org/abs},
      pages={20},
      year={2009}
}

@inproceedings{gema2025we,
  title={Are we done with mmlu?},
  author={Gema, Aryo Pradipta and Leang, Joshua Ong Jun and Hong, Giwon and Devoto, Alessio and Mancino, Alberto Carlo Maria and Saxena, Rohit and He, Xuanli and Zhao, Yu and Du, Xiaotang and Madani, Mohammad Reza Ghasemi and others},
  booktitle={Proceedings of the 2025 Conference of the Nations of the Americas Chapter of the Association for Computational Linguistics: Human Language Technologies (Volume 1: Long Papers)},
  pages={5069--5096},
  year={2025}
}

@article{wang2024mmlu,
  title={Mmlu-pro: A more robust and challenging multi-task language understanding benchmark},
  author={Wang, Yubo and Ma, Xueguang and Zhang, Ge and Ni, Yuansheng and Chandra, Abhranil and Guo, Shiguang and Ren, Weiming and Arulraj, Aaran and He, Xuan and Jiang, Ziyan and others},
  journal={Advances in Neural Information Processing Systems},
  volume={37},
  pages={95266--95290},
  year={2024}
}

@inproceedings{suzgun2023challenging,
  title={Challenging big-bench tasks and whether chain-of-thought can solve them},
  author={Suzgun, Mirac and Scales, Nathan and Sch{\"a}rli, Nathanael and Gehrmann, Sebastian and Tay, Yi and Chung, Hyung Won and Chowdhery, Aakanksha and Le, Quoc and Chi, Ed and Zhou, Denny and others},
  booktitle={Findings of the Association for Computational Linguistics: ACL 2023},
  pages={13003--13051},
  year={2023}
}

@article{du2025supergpqa,
  title={Supergpqa: Scaling llm evaluation across 285 graduate disciplines},
  author={Du, Xinrun and Yao, Yifan and Ma, Kaijing and Wang, Bingli and Zheng, Tianyu and Zhu, King and Liu, Minghao and Liang, Yiming and Jin, Xiaolong and Wei, Zhenlin and others},
  journal={arXiv preprint arXiv:2502.14739},
  year={2025}
}

@inproceedings{rein2024gpqa,
  title={Gpqa: A graduate-level google-proof q\&a benchmark},
  author={Rein, David and Hou, Betty Li and Stickland, Asa Cooper and Petty, Jackson and Pang, Richard Yuanzhe and Dirani, Julien and Michael, Julian and Bowman, Samuel R},
  booktitle={First Conference on Language Modeling},
  year={2024}
}

@article{cobbe2021training,
  title={Training verifiers to solve math word problems},
  author={Cobbe, Karl and Kosaraju, Vineet and Bavarian, Mohammad and Chen, Mark and Jun, Heewoo and Kaiser, Lukasz and Plappert, Matthias and Tworek, Jerry and Hilton, Jacob and Nakano, Reiichiro and others},
  journal={arXiv preprint arXiv:2110.14168},
  year={2021}
}

@article{hendrycks2024measuring,
  title={Measuring mathematical problem solving with the math dataset, 2021},
  author={Hendrycks, Dan and Burns, Collin and Kadavath, Saurav and Arora, Akul and Basart, Steven and Tang, Eric and Song, Dawn and Steinhardt, Jacob},
  journal={URL https://arxiv. org/abs/2103.03874},
  volume={2},
  year={2024}
}

@article{liu2023your,
  title={Is your code generated by chatgpt really correct? rigorous evaluation of large language models for code generation},
  author={Liu, Jiawei and Xia, Chunqiu Steven and Wang, Yuyao and Zhang, Lingming},
  journal={Advances in Neural Information Processing Systems},
  volume={36},
  pages={21558--21572},
  year={2023}
}

@article{chen2021evaluating,
  title={Evaluating large language models trained on code},
  author={Chen, Mark},
  journal={arXiv preprint arXiv:2107.03374},
  year={2021}
}

@article{austin2021program,
  title={Program synthesis with large language models},
  author={Austin, Jacob and Odena, Augustus and Nye, Maxwell and Bosma, Maarten and Michalewski, Henryk and Dohan, David and Jiang, Ellen and Cai, Carrie and Terry, Michael and Le, Quoc and others},
  journal={arXiv preprint arXiv:2108.07732},
  year={2021}
}

@article{gu2024cruxeval,
  title={Cruxeval: A benchmark for code reasoning, understanding and execution},
  author={Gu, Alex and Rozi{\`e}re, Baptiste and Leather, Hugh and Solar-Lezama, Armando and Synnaeve, Gabriel and Wang, Sida I},
  journal={arXiv preprint arXiv:2401.03065},
  year={2024}
}

@article{zellers2019hellaswag,
  title={Hellaswag: Can a machine really finish your sentence?},
  author={Zellers, Rowan and Holtzman, Ari and Bisk, Yonatan and Farhadi, Ali and Choi, Yejin},
  journal={arXiv preprint arXiv:1905.07830},
  year={2019}
}

@article{clark2019boolq,
  title={Boolq: Exploring the surprising difficulty of natural yes/no questions},
  author={Clark, Christopher and Lee, Kenton and Chang, Ming-Wei and Kwiatkowski, Tom and Collins, Michael and Toutanova, Kristina},
  journal={arXiv preprint arXiv:1905.10044},
  year={2019}
}

@inproceedings{bisk2020piqa,
  title={Piqa: Reasoning about physical commonsense in natural language},
  author={Bisk, Yonatan and Zellers, Rowan and Gao, Jianfeng and Choi, Yejin and others},
  booktitle={Proceedings of the AAAI conference on artificial intelligence},
  volume={34},
  number={05},
  pages={7432--7439},
  year={2020}
}

@article{sap2019socialiqa,
  title={Socialiqa: Commonsense reasoning about social interactions},
  author={Sap, Maarten and Rashkin, Hannah and Chen, Derek and LeBras, Ronan and Choi, Yejin},
  journal={arXiv preprint arXiv:1904.09728},
  year={2019}
}

@article{kwiatkowski2019natural,
  title={Natural questions: a benchmark for question answering research},
  author={Kwiatkowski, Tom and Palomaki, Jennimaria and Redfield, Olivia and Collins, Michael and Parikh, Ankur and Alberti, Chris and Epstein, Danielle and Polosukhin, Illia and Devlin, Jacob and Lee, Kenton and others},
  journal={Transactions of the Association for Computational Linguistics},
  volume={7},
  pages={453--466},
  year={2019},
  publisher={MIT Press One Rogers Street, Cambridge, MA 02142-1209, USA journals-info~…}
}

@article{clark2018think,
      title={Think you have solved question answering? try arc, the ai2 reasoning challenge},
      author={Clark, Peter and Cowhey, Isaac and Etzioni, Oren and Khot, Tushar and Sabharwal, Ashish and Schoenick, Carissa and Tafjord, Oyvind},
      journal={arXiv preprint arXiv:1803.05457},
      year={2018}
}

@article{sakaguchi2021winogrande,
  title={Winogrande: An adversarial winograd schema challenge at scale},
  author={Sakaguchi, Keisuke and Bras, Ronan Le and Bhagavatula, Chandra and Choi, Yejin},
  journal={Communications of the ACM},
  volume={64},
  number={9},
  pages={99--106},
  year={2021},
  publisher={ACM New York, NY, USA}
}

@article{dua2019drop,
  title={DROP: A reading comprehension benchmark requiring discrete reasoning over paragraphs},
  author={Dua, Dheeru and Wang, Yizhong and Dasigi, Pradeep and Stanovsky, Gabriel and Singh, Sameer and Gardner, Matt},
  journal={arXiv preprint arXiv:1903.00161},
  year={2019}
}

@article{joshi2017triviaqa,
  title={Triviaqa: A large scale distantly supervised challenge dataset for reading comprehension},
  author={Joshi, Mandar and Choi, Eunsol and Weld, Daniel S and Zettlemoyer, Luke},
  journal={arXiv preprint arXiv:1705.03551},
  year={2017}
}

@article{ifeval2024,
title={IFEval: Instruction-Following Evaluation for Large Language Models},
author={Liu, Zhengxiao and others},
journal={arXiv preprint arXiv:2404.01927},
year={2024}
}

@misc{AIME25,
  author = {AIME},
  title = {{AIME problems and solutions}},
  howpublished = {\url{https://artofproblemsolving.com/wiki/index.php/AIME_Problems_and_Solutions.}},
  year = {2025}
}

@article{white2024livebench,
  title={Livebench: A challenging, contamination-free llm benchmark},
  author={White, Colin and Dooley, Samuel and Roberts, Manley and Pal, Arka and Feuer, Ben and Jain, Siddhartha and Shwartz-Ziv, Ravid and Jain, Neel and Saifullah, Khalid and Naidu, Siddartha and others},
  journal={arXiv preprint arXiv:2406.19314},
  volume={4},
  year={2024}
}

@inproceedings{lightman2023let,
  title={Let's verify step by step},
  author={Lightman, Hunter and Kosaraju, Vineet and Burda, Yuri and Edwards, Harrison and Baker, Bowen and Lee, Teddy and Leike, Jan and Schulman, John and Sutskever, Ilya and Cobbe, Karl},
  booktitle={The Twelfth International Conference on Learning Representations},
  year={2023}
}

@article{lin2025zebralogic,
  title={Zebralogic: On the scaling limits of llms for logical reasoning},
  author={Lin, Bill Yuchen and Bras, Ronan Le and Richardson, Kyle and Sabharwal, Ashish and Poovendran, Radha and Clark, Peter and Choi, Yejin},
  journal={arXiv preprint arXiv:2502.01100},
  year={2025}
}

@article{xie2023doremi,
  title={Doremi: Optimizing data mixtures speeds up language model pretraining},
  author={Xie, Sang Michael and Pham, Hieu and Dong, Xuanyi and Du, Nan and Liu, Hanxiao and Lu, Yifeng and Liang, Percy S and Le, Quoc V and Ma, Tengyu and Yu, Adams Wei},
  journal={Advances in Neural Information Processing Systems},
  volume={36},
  pages={69798--69818},
  year={2023}
}

@article{jain2024livecodebench,
  title={Livecodebench: Holistic and contamination free evaluation of large language models for code},
  author={Jain, Naman and Han, King and Gu, Alex and Li, Wen-Ding and Yan, Fanjia and Zhang, Tianjun and Wang, Sida and Solar-Lezama, Armando and Sen, Koushik and Stoica, Ion},
  journal={arXiv preprint arXiv:2403.07974},
  year={2024}
}

\clearpage

\section{Appendix}
\label{sec:appendix}

\subsection{Contributions}

All authors are alphabetically sorted by last name.

\textbf{Technical and management leadership}: Sungmin Lee, Junghwan Lim

\textbf{Core contributors}: Dongseok Kim, Taehyun Kim, Eunhwan Park, Jeesoo Lee, Jeongdoo Lee, Junhyeok Lee

\textbf{Contributors}: Wai Ting Cheung, Dahye Choi, Jaeheui Her, Jaeyeon Huh, Hanbin Jung, Changjin Kang, Beomgyu Kim, Minjae Kim, Taewhan Kim, Youngrok Kim, Hyukjin Kweon, Haesol Lee, Kungyu Lee, Dongpin Oh, Yeongjae Park, Bokki Ryu, Dongjoo Weon

\end{document}